\documentclass[conference]{IEEEtran}
\IEEEoverridecommandlockouts
\usepackage{cite}
\usepackage{amsmath,amssymb,amsfonts}
\usepackage{algorithm,algorithmic}
\usepackage{graphics,graphicx} 
\usepackage{subfigure}
\usepackage{textcomp}
\usepackage{xcolor}
\usepackage{float}
\usepackage{calrsfs}
\def\BibTeX{{\rm B\kern-.05em{\sc i\kern-.025em b}\kern-.08em
		T\kern-.1667em\lower.7ex\hbox{E}\kern-.125emX}}
\begin{document}
	
	\title{EGFC: Evolving Gaussian Fuzzy Classifier from Never-Ending Semi-Supervised Data Streams -- With Application to Power Quality Disturbance Detection and Classification}
		
	\author{
		\IEEEauthorblockN{Daniel Leite}
		\IEEEauthorblockA{\textit{Federal University of Lavras}\\ Department of Automatics\\
			Lavras, Brazil \\
			daniel.leite@ufla.br}
		\and	
		\IEEEauthorblockN{Leticia Decker}
		\IEEEauthorblockA{\textit{University of Bologna}\\
			INFN Bologna \\
			Bologna, Italy \\
			leticia.deckerde@unibo.it}
		\and
		\IEEEauthorblockN{Marcio Santana, and Paulo Souza}
		\IEEEauthorblockA{\textit{Federal Center for Technological Education}\\
			CEFET-MG \\
			Minas Gerais, Brazil \\
			\{marciosantana, paulovitor\}@cefetmg.br}
	}
	
	\maketitle

\begin{abstract}

Power-quality disturbances lead to several drawbacks such as limitation of the production capacity, increased line and equipment currents, and consequent ohmic losses; higher operating temperatures, premature faults, reduction of life expectancy of machines, malfunction of equipment, and unplanned outages. Real-time detection and classification of disturbances are deemed essential to industry standards. We propose an Evolving Gaussian Fuzzy Classification (EGFC) framework for semi-supervised disturbance detection and classification combined with a hybrid Hodrick-Prescott and Discrete-Fourier-Transform attribute-extraction method applied over a landmark window of voltage waveforms. Disturbances such as spikes, notching, harmonics, and oscillatory transient are considered. Different from other monitoring systems, which require offline training of models based on a limited amount of data and occurrences, the proposed online data-stream-based EGFC method is able to learn disturbance patterns autonomously from never-ending data streams by adapting the parameters and structure of a fuzzy rule base on the fly. Moreover, the fuzzy model obtained is linguistically interpretable, which improves model acceptability. We show encouraging classification results.

\end{abstract}

\section{Introduction}

Power system disturbance detection and classification are broad and difficult research issues. Detection and classification are generally handled by using machine learning and computational intelligence methods \cite{Achlerkar,Li,Sobrinho}. The number of correlated attributes potentially carrying important information about disturbances is high. As real-time monitoring of all attributes is infeasible, a method to extract the most prominent ones to assist detection is fundamental. A further issue concerns the frequent occurrence of new situations. The emergence of new patterns in power system data affects the performance of classification systems. Novelties may arise since the involved signals are mutually related and time-varying. Moreover, the superposition of different events in different intensities may generate never-before-seen data, which tends to confuse conventional offline-trained classification systems \cite{Skrjanc,LeiteD1}.

Online disturbance-detection methods deal with the occurrence of new patterns in data streams. New behaviors should be captured by adaptive or evolving models, namely, models supplied with incremental learning algorithms \cite{Andonovski,Hu,Silva,Souza}. Generally, the amount of data in power applications is large. Therefore, storing the data for further offline development of equation-based or intelligent models, and statistical analyses, is quite often time-consuming, useless, or even impossible. Adaptive and evolving modeling from online data streams are distinct concepts. \textit{Adaptive models} (parametrically adaptive models from the control theory) are suitable to cope with smooth, gradual changes of system parameters and statistical properties of data (concept drift). However, when an adaptive model is changed to learn a new behavior, the knowledge about some previous behaviors tends to be partially lost. This phenomenon is known as \textit{catastrophic forgetting} \cite{Kirkpatrick}. Abrupt changes of the values of a variable or parameter (concept shift) require both parametrical and structural adaptation of models. Such higher level of flexibility of models, which includes the incremental update of the model structure, outlines a broad research area known as \textit{evolving intelligence} \cite{AngelovP,LughoferE,LeiteD2,Cordovil}.

Evolving fuzzy rule-based models are appropriate for online detection and classification in nonstationary data stream environments, such as that found in power systems \cite{LeiteD1,Andonovski,Silva}. Incremental fuzzy clustering algorithms have been used for constructing rule-based evolving classifiers. These algorithms are capable of determining the classifier structure and parameters from scratch based on online data. Development and use of rule-based evolving systems have grown in the last decade. Successful applications of these systems in complex real-world problems, including control, prediction, classification, identification, and function approximation, are found  \cite{Skrjanc,AngelovP,LughoferE,LeiteD2,Cordovil,Petronio}. A further advantage of evolving fuzzy systems is that they may provide linguistically-appealing granular information \cite{LeiteD2,Cordovil,PedryczW}, that is, these systems may explain their results or actions. Online structural adaptation of a fuzzy model to handle nonstationarities is pursued by adding, merging, and removing rules from a knowledge base \cite{Skrjanc}.

This paper addresses a new fuzzy modeling framework we called Evolving Gaussian Fuzzy Classification (EGFC) framework. EGFC is a semi-supervised variation of evolving granular rule-based approach \cite{LeiteD2,LeiteD3} for the construction of nonlinear and time-varying classifiers -- being unsupervised and supervised learning the boundary cases. We aim to detect and classify anomalies, broadly speaking, and power quality disturbances as a particular application example. An EGFC model employs Gaussian membership functions to associate numerical input data with classes. Its incremental learning algorithm provides a dynamic classifier with simple math and linguistic rules describing its decisions. The set of EGFC rules represents the essence of a data stream. From a point of view, the EGFC approach consists in looking to stream data, and deciding between coarser or more detailed granules to achieve a better classification accuracy, and provide decision making support. Spikes, notching, harmonics, and oscillatory-transient types of disturbances in power systems are taken into consideration in the present study. 

A hybrid method to extract highly discriminative attributes to be used as inputs of the EGFC model is also addressed. The method combines a Hodrick-Prescott (HP) filter \cite{Hodrick} and Discrete Fourier Transform (DFT) \cite{Lathi} applied over a landmark window of voltage data to provide attributes that help disturbance classification. In particular, the method is different from those addressed in related power-system literature as HP analysis provides a smooth nonlinear representation that is sensitive to long and short-term changes. In other words, a series of smooth nonlinear trends are obtained, whereas the information removed from the original data is maintained separately and can be accessed for analysis. In comparison to S-transform-based attribute extraction methods \cite{Li,Reddy}, often used in related literature, the proposed hybrid HP-DFT method has proven to be faster and effective -- important characteristics for high-frequency data-stream processing.

The rest of this paper is organized as follows. Section \ref{feature} describes the HP and DFT methods for attribute extraction. Section \ref{evolving} presents the data-stream-based semi-supervised learning algorithm and the Evolving Gaussian Fuzzy Classifier, EGFC, proposed for a broad class of classification and online anomaly-detection problems from numerical data. Section \ref{methodology} describes the methodology for generating the data and developing the classifier. Results are given in Section \ref{experimental}. The conclusion is outlined in Section \ref{conclusion}.

\section{Attribute Extraction}\label{feature}

We describe an approach to extract attributes that indicate the occurrence of disturbances in voltage data. Fundamentally, an HP filter and the DFT are applied to raw voltage data within a time window. A more discriminative set of attributes facilitates model interpretation, reduces data overfitting, and may produce better results due to the elimination of attributes and noise that may mislead online systems \cite{PedryczW,LeiteD2}.

\subsection{Hodrick-Prescott Filter}

An HP filter decomposes a signal, $y_t$, into its trend, $\tau_t$, and cyclical and random components, $C_t$, such that $y_t = \tau_t + C_t$. It is equivalent to a cubic spline smoother, with the smoothed portion in $\tau_t$ \cite{Hodrick}. In essence, low-frequency fluctuations are separated from the original data. The separation hypothesis is that the low-frequency variability represents the long-term trend, whereas the high-frequency variability means random phenomena. The HP filter is for the first time used for power system disturbance classification in this paper.

The HP filter extracts the trend, which is stochastic, but with smooth variations over time that are uncorrelated with other variations. The idea is to minimize the functional

\vspace{-5pt}

\begin{eqnarray}
J = \sum_{t=1}^T C_t^2 + \lambda \sum_{t=2}^{T-1} (\Delta^2 \tau_t)^2, \label{eq1}
\end{eqnarray}

\noindent with respect to $\tau_t$, in which $C_t := y_t - \tau_t$, and $y_t$, $t = 1,2,...,T$, denotes the underlying signal; $\Delta^2 := (1-L)^2$; $L$ is the lag operator, e.g., $Lx_t = x_{t-1}$; $T$ is the amount of data samples; and $\lambda$ penalizes the variability of the trend component. Parameter $\lambda$ is the smoothing parameter; it controls the variation of the growth rate of the trend. The first term of \eqref{eq1} is the sum of deviations of the signal concerning the square trend, a measure of the degree of fit. The second term is the sum of squares of the second differences of the trend component, a measure of the degree of smoothness. See \cite{Hodrick} for details. The fourth section of this paper provides practical examples of the HP decomposition specifically for disturbances detection.

\subsection{Discrete Fourier Transform}

The Fourier Transform is one of the most used frequency-domain signal processing tools. The idea is that any periodic signal can be described by a sum of sines and cosines \cite{Lathi}. When measured data are available, a frequency spectrum is generated by discrete Fourier transformation from

\vspace{-5pt}

\begin{eqnarray}
DFT(f_n) = \frac {1}{N} \sum_{k=0}^{N-1} x_k e^{-j2\pi f_n k\Delta t},
\end{eqnarray}

\noindent in which $x_k$ is a discrete signal; $\Delta t$ corresponds to time intervals; and $N:=T/\Delta t$ is the number of samples -- being $T$ the total time interval. Additionally, $f_n = n/T$, $n = 0, 1, ..., N-1$, are the frequency components.

The discrete Fourier transform is an invertible linear transformation. The inverse is given by

\vspace{-4pt}

\begin{eqnarray}
x_k = \frac {1}{\Delta t} \sum_{{f_n}=0}^{(1-N)/T} DFT(f_n) e^{j2\pi f_n k\Delta t}.
\end{eqnarray}

\noindent See \cite{Lathi} for details on discrete Fourier transforms.

\subsection{Root Mean Square Voltage}

The effective (RMS) value is a measure of the magnitude of a variable quantity. RMS values can be calculated for a sequence of discrete values. The effective voltage of an alternating current circuit may provide evidence of some types of disturbances, e.g., sag, swell, and interruptions. 

The RMS value of a sinusoidal waveform $\textbf{x}$ for a set of $N$ samples, $\textbf{x} = [x_1, x_2, ..., x_N]$, is

\begin{eqnarray}
x_{RMS} = \sqrt {\frac {1}{N} \sum_{k=1}^{N} x_k^2}.
\end{eqnarray}

\noindent This is an equivalent direct value able to produce the same power as that of the original waveform.

\section{Evolving Gaussian Fuzzy Classifier from Never-Ending Semi-Supervised Data Streams}\label{evolving}

\subsection{Preliminaries}
	
We present EGFC, a semi-supervised evolving classifier derived from the online granular-computing framework of Leite et al. \cite{LeiteD2,LeiteD3}. EGFC employs Gaussian membership functions to cover the data space with fuzzy granules (local models), and associate new numerical data to class labels. Granules are scattered in the data space wherever needed to represent local information. EGFC overall response comes from the fuzzy aggregation of local models. A recursive algorithm constructs its rule base, and updates local models to deal with novelties. EGFC addresses issues such as unlimited amounts of data and scalability \cite{LeiteD2}.

Local EGFC models are created if the new data are sufficiently different from the current knowledge. The learning algorithm can expand, retract, delete, and merge granules on occasion. Rules are reviewed according to inter-granular relations. EGFC provides nonlinear, nonstationary, and smooth boundaries among classes. This paper particularly addresses a 5-class disturbance classification problem.

Formally, let an input-output pair $(\textbf{x}, y)$ be related through $y = f(\textbf{x})$. We seek an approximation to $f$ to estimate the value of $y$ given $\textbf{x}$. In classification, $y$ is a class label, a value in a set $\{C_1, ..., C_m\} \in \mathbb{N}^m $, and $f$ specifies class boundaries. In the more general, semi-supervised case, $C_k$ may or may not be known when $\textbf{x}$ arrives. Classification of never-ending data streams involves pairs $(\textbf{x}, C)^{[h]}$ of time-sequenced data, indexed by $h$. Nonstationarity requires evolving classifiers to identify time-varying relations $f^{[h]}$.

\subsection{Gaussian Functions and Rule Structure}

Learning in EGFC does not require initial rules. Rules are created and dynamically updated depending on the behavior of a system over time. When a data sample is available, a decision procedure may add a rule to the model structure or update the parameters of a chosen rule.

In EGFC models, a rule $R^i$ is

\vspace{4pt}

~~~~ IF $(x_1 $ is $ A_1^i)$ AND ... AND $(x_n $ is $ A_n^i)$

~~~~ THEN $(y $ is $ C^i)$ 

\vspace{5pt}

\noindent in which $x_j$, $j = 1, ..., n$, are input attributes, and $y$ is the output (a class). The data stream is denoted ${(\textbf{x}, y)}^{[h]}, h = 1, ...$ Moreover, $A_j^i$,  $j = 1, ..., n$; $i = 1, ..., c$, are Gaussian membership functions built from the available data; and $C^i$ is the class label of the $i$-th rule. Rules $R^i$, $i = 1, ..., c$, form the rule base. The number of rules, $c$, is variable, which is a notable characteristic of the approach since guesses on how many data partitions exist are needless \cite{Skrjanc,LeiteD2}.

A normal Gaussian function, $A_j^i = G(\mu_j^i, \sigma_j^i)$, has height 1 \cite{PedryczW}. It is characterized by the modal value $\mu_j^i$ and dispersion $\sigma_j^i$. Characteristics that make Gaussians appropriate include: (i) easiness of learning and changing, i.e., modal values and dispersions are updated straightforwardly from a data stream; (ii) infinite support, i.e., since the data are priorly unknown, the support of Gaussians extends to the whole domain; and (iii) smooth surface of fuzzy granules, $\gamma^i = A^i_1 \times ... \times A^i_j \times ... \times A^i_n$, in the $n$-dimensional Cartesian space -- obtained by the cylindrical extension of one-dimensional Gaussians, and the use of the minimum T-norm aggregation \cite{PedryczW}.

\subsection{Adding Rules to the Evolving Fuzzy Classifier}

Rules may not exist \textit{a priori}. They are created and evolved as data are available. A new granule $\gamma^{c+1}$ and the rule $R^{c+1}$ that governs the granule are created if none of the existing rules $\{R^1, ..., R^c\}$ are sufficiently activated by $\textbf{x}^{[h]}$; i.e., $\textbf{x}^{[h]}$ brings new information. Let $\rho^{[h]} \in [0,1]$ be an adaptive threshold that determines if a new rule is needed. If

\begin{eqnarray}
T\left(A_1^i(x_1^{[h]}),...,A_n^i(x_n^{[h]})\right)\leq \rho^{[h]}, ~ \forall i, ~ i = 1, ..., c, \label{activ}
\end{eqnarray}

\noindent in which $T$ is any triangular norm, then the EGFC structure is expanded. The minimum (G{\"o}del) T-norm is used in this paper. If $\rho^{[h]}$ is equal to 0, then the model is structurally stable, and unable to follow concept shifts. In contrast, if $\rho^{[h]}$ is equal to 1, EGFC creates a rule for each new sample, which is not practical. Structural and parametric adaptability are balanced for intermediate values (stability-plasticity tradeoff) \cite{LeiteD4}.

The value of $\rho^{[h]}$ is crucial to regulate how large granules can be. Different choices impact the accuracy and compactness of a model, resulting in different granular perspectives of the same problem. Section III-E gives a Gaussian-dispersion-based procedure to update $\rho^{[h]}$.

A new granule $\gamma^{c+1}$ is initially represented by membership functions, $A_j^{c + 1}$, $j = 1, ..., n$, with

\vspace{-6pt}

\begin{eqnarray}
\mu_j^{c+1} = x_j^{[h]}, \label{eq6}
\end{eqnarray}

\noindent and 
 
\vspace{-6pt}
 
\begin{eqnarray}
\sigma_j^{c+1} = 1/2\pi. \label{eq7}
\end{eqnarray}

\noindent We call \eqref{eq7} the Stigler approach to standard Gaussian functions, or \textit{maximum approach} \cite{Stigler,LeiteD5}. The intuition is to start big, and let the dispersions gradually shrink when new samples activate the same granule. This strategy is appealing for a compact model structure.

In general, the class $C^{c+1}$ of the rule $R^{c+1}$  is initially undefined, i.e., the $(c+1)$-th rule remains unlabeled until a label is provided. If the corresponding output, $y^{[h]}$, associated to $\textbf{x}^{[h]}$, becomes available, then

\vspace{-4pt}

\begin{eqnarray}
C^{c+1} = y^{[h]}. \label{eq8}
\end{eqnarray}

\noindent Otherwise, the first labeled sample that arrives after the $h$-th time step, and activates the rule $R^{c+1}$ according to \eqref{activ}, is used to define its class, $C^{c+1}$.

In case a labeled sample activates a rule that is already labeled, but their labels are different, then a new (partially overlapped) granule and a rule are created to represent new information. Partially overlapped Gaussian granules tagged with different labels tend to have their dispersions reduced over time by the parameter adaptation procedure (Sec. III-D). The modal values of the Gaussian granules may also drift, if convenient for a more suitable decision boundary.

With this initial parameterization, preference is given to granules balanced along their dimensions, rather than granules with unbalanced geometry. EGFC realizes the principle of the balanced granularity \cite {Wang}, but allows the Gaussians to find more appropriate places and dispersions.

\subsection{Incremental Parameter Updating}

Updating the EGFC model consists in: (i) reducing or expanding Gaussians $A_j^{i^*}$, $j = 1, ..., n$, of the most active granule, $\gamma^{i^*}$, considering labeled and unlabeled samples; (ii) moving granules toward regions of relatively dense population; and (iii) tagging rules when labeled data are available. Adaptation aims to develop more specific local models  \cite{Yager}, and provide pavement (covering) to new data.

A rule $R^i$ is candidate to be updated if it is sufficiently activated by an unlabeled sample, $\textbf{x}^{[h]}$, according to

 \begin{eqnarray}
min\left(A_1^i(x_1^{[h]}),...,A_n^i(x_n^{[h]})\right) > \rho^{[h]}. \label{eq9}
\end{eqnarray} 

\noindent Geometrically, $\textbf{x}^{[h]}$ belongs to a region highly influenced by the granule $\gamma^i$. Only the most active rule, $R^{i^*}$, is chosen for adaptation in case two or more rules reach the $\rho^{[h]}$ level for the unlabeled $\textbf{x}^{[h]}$. For a labeled sample, i.e., for pairs $(\textbf{x},y)^{[h]}$, the class of the most active rule $R^{i^*}$, if defined, must match $y^{[h]}$. Otherwise, the second most active rule among those that reached the $\rho^{[h]}$ level is chosen for adaptation, and so on. If none of the rules are apt, a new one is created (Sec. III-C).

To include $\textbf{x}^{[h]}$ in $R^{i^*}$, EGFC's learning algorithm updates the modal values and dispersions of the corresponding membership functions $A_j^{i^*}$, $j = 1, ..., n$, from

\begin{eqnarray}
\mu_j^{i^*}(new) = \frac {(\varpi^{i^*}-1) \mu_j^{i^*}(old) + x_j^{[h]}}{\varpi^{i^*}}, \label{eq10}
\end{eqnarray} 

\noindent and

\begin{eqnarray}
\sigma_j^{i^*}(new) &=& \biggl( \frac {(\varpi^{i^*}-1)}{\varpi^{i^*}} ~ \left(\sigma_j^{i^*}(old)\right)^2 ~+ \nonumber \\
&& +~ \frac {1}{\varpi^{i^*}} \left(x_j^{[h]} - \mu_j^{i^*}(old)\right)^2 \biggr)^{1/2}, ~~~ \label{eq11}
\end{eqnarray} 

\noindent in which $\varpi^{i^*}$ is the number of times the $i^*-th$ rule was chosen to be updated. Notice that \eqref{eq10}-\eqref{eq11} are recursive and, therefore, do not require data storage. As $\sigma^{i^*}$ defines a convex region of influence around $\mu^{i^*}$, very large and very small values may induce, respectively, a unique or too many granules per class. An approach is to keep $\sigma_j^{i^*}$ between a lower, $1/4\pi$, and the Stigler, $1/2\pi$, limits.

\subsection{Dispersion-Based Time-Varying $\rho$-Level}

Let the activation threshold, $\rho^{[h]} \in [0,1]$, be time-varying. The threshold assumes values in the unit interval according to the overall average dispersion

\vspace{-8pt}

\begin{eqnarray}
\sigma_{avg}^{[h]} &=& \frac{1}{cn} \sum\limits_{i=1}^c \sum\limits_{j=1}^n \sigma^{i[h]}_j; \label{eq12}
\end{eqnarray} 

\noindent $c$ and $n$ are the number of rules and attributes, so that

\vspace{-8pt}

\begin{eqnarray}
\rho(new) &=& \frac{\sigma_{avg}^{[h]}}{\sigma_{avg}^{[h-1]}} ~ \rho(old). \label{eq13}
\end{eqnarray} 

As mentioned, rules' activation levels for an input $\textbf{x}^{[h]}$ are compared to $\rho^{[h]}$ to decide between parametric or structural changes of an EGFC model. In general, EGFC starts learning from an empty rule base, and without knowledge about the properties of the data. Practice suggests $\rho^{[0]} = 0.1$ as starting value. The threshold tends to converge to a proper value if the classifier structure achieves a level of maturity and stability. Nonstationarities and new classes guide $\rho^{[h]}$ to values that better reflect the needs of the current environment.

\subsection{Merging Similar Granules}

Similarity between two granules with the same class label may be high enough to form a unique granule that inherits the essence of both. Analysis of inter-granular relations requires a distance measure between Gaussians. Let

\vspace{-8pt}

\begin{eqnarray}
d(\gamma^{i_1}, \gamma^{i_2}) &=& \frac {1}{n} \biggl( ~\sum_{j=1}^n | \mu_j^{i_1} - \mu_j^{i_2} | + \sigma_j^{i_1} +  \nonumber \\ && \sigma_j^{i_2} - 2 \sqrt{\sigma_j^{i_1} \sigma_j^{i_2}} ~\biggr)  \label{eq14}
\end{eqnarray}

\noindent be the distance between $\gamma^{i_1}$ and $\gamma^{i_2}$. This measure considers the information specificity, that is, in turn, inversely related to the Gaussians' dispersion \cite{LeiteD5}. For example, if the dispersions $\sigma_j^{i_1}$ and $\sigma_j^{i_2}$ differ one from another, rather than being equal, the distance between the underlying Gaussians is larger.

EGFC may merge the pair of granules that presents the smallest value of $d(.)$ for all pairs of granules. Both granules must be either unlabeled or tagged with the same class label. The merging decision is based on a threshold value, $\Delta$, or expert judgment regarding the suitability of combining such granules to have a more compact model. For data within the unit hypercube, we suggest $\Delta = 0.1$ as default, which means that the candidate granules should be quite similar.

A new granule, say $\gamma^i$, which results from $\gamma^{i_1}$ and $\gamma^{i_2}$, is built by Gaussians with modal values

\vspace{-8pt}

\begin{eqnarray}
\mu_j^i = \frac{\frac{\sigma_j^{i_1}}{\sigma_j^{i_2}}\mu_j^{i_1} + \frac{\sigma_j^{i_2}}{\sigma_j^{i_1}}\mu_j^{i_2}}{\frac{\sigma_j^{i_1}}{\sigma_j^{i_2}} + \frac{\sigma_j^{i_2}}{\sigma_j^{i_1}}}, ~  j=1,...,n,  \label{eq15}
\end{eqnarray}

\noindent and dispersion

\vspace{-8pt}

\begin{eqnarray}
\sigma_j^i = \sigma_j^{i_1} + \sigma_j^{i_2}, ~ j=1,...,n. \label{eq16}
\end{eqnarray}

\noindent These relations take into account the granular uncertainty to find an appropriate location and size to the resulting granule. Merging minimizes redundancy \cite{Skrjanc,LeiteD2}.

\subsection{Deleting Rules}

A rule is removed from the EGFC model if it is inconsistent with the current environment. In other words, if a rule is not activated for a number of iterations, say $h_r$, then it is deleted from the rule base. However, if a class is rare, e.g., a type of power quality disturbance is unusual, then it may be the case to set $h_r$ to infinity and keep the inactive rules. Removing rules periodically helps to keep the model updated.

\subsection{Semi-Supervised Learning from Data Streams}

The semi-supervised learning procedure to construct and update EGFC models along their lifespan is given below.

~~

\hrule
\vspace{5pt}
\textbf{EGFC: Online Semi-Supervised Learning}
\vspace{3pt}
\hrule
\vspace{3pt}
\begin{algorithmic}[1]
	\STATE Initial number of rules, $c = 0$;
	\STATE Initial meta-parameters, $\rho^{[0]} = \Delta = 0.1$, $h_r = 200$;
    \STATE Read input data sample $\textbf{x}^{[h]}, h=1$;
    \STATE Create granule $\gamma^{c+1}$ (Eqs. \eqref{eq6}-\eqref{eq7}), unknown class $C^{c+1}$;
    \STATE \textbf{FOR} $h$ = 2, ... \textbf{DO}
    \STATE ~~ Read $\textbf{x}^{[h]}$, calculate rules' activation degree (Eq. \eqref{activ});
    \STATE ~~ Determine the most active rule $R^{i^*}$;
    \STATE ~~ Provide estimated class $C^{i^*}$;
    \STATE ~~ // Model adaptation
    \STATE ~~ \textbf{IF} $T(A^i_1(x_1^{[h]}), ..., A^i_n(x_n^{[h]})) \leq \rho^{[h]} ~ \forall i, ~ i = 1, ..., c$
    \STATE ~~~~ \textbf{IF} actual label $y^{[h]}$ is available
    \STATE ~~~~~~ Create labeled granule $\gamma^{c+1}$ (Eqs. \eqref{eq6}-\eqref{eq8});
    \STATE ~~~~ \textbf{ELSE}
    \STATE ~~~~~~ Create unlabeled granule $\gamma^{c+1}$ (Eqs. \eqref{eq6}-\eqref{eq7});
    \STATE ~~~~ \textbf{END}
    \STATE ~~ \textbf{ELSE}
    \STATE ~~~~ \textbf{IF} actual label $y^{[h]}$ is available
    \STATE ~~~~~~ Update the most active granule $\gamma^{i^*}$ whose class \\ 
    ~~~~~~ $C^{i^*}$ \hspace{-6pt} is equal to $y^{[h]}$ (Eqs. \eqref{eq10}-\eqref{eq11});
    \STATE ~~~~~~ Tag unlabeled active granules;
    \STATE ~~~~ \textbf{ELSE}
    \STATE ~~~~~~ Update the most active  $\gamma^{i^*}$ (Eqs. \eqref{eq10}-\eqref{eq11});
    \STATE ~~~~ \textbf{END}
    \STATE ~~ \textbf{END}
    \STATE ~~ Update the $\rho$-level (Eqs. \eqref{eq12}-\eqref{eq13});
    \STATE ~~ Delete inactive rules based on $h_r$;
    \STATE ~~ Merge granules based on $\Delta$ (Eqs. \eqref{eq14}-\eqref{eq16});

    \STATE \textbf{END}    
\end{algorithmic}
\vspace{3pt}
\hrule
\vspace{5pt}

~~

\vspace{-5pt}

\section{Methodology}\label{methodology}

We describe the methodology to generate power system disturbances. We give examples of disturbances, and a flowchart that connects the DFT-HP attribute extraction and EGFC.

\subsection{Online Monitoring System}

Voltage data from a 13.8kV grid are produced according to the IEEE standard \cite{std1159}. The fundamental and sampling frequencies are 60Hz and 15,360Hz. Thus, 256 samples per cycle are given. This sampling rate is sufficient to characterize most of the disturbances in power systems, including spikes, notching, harmonics, and oscillatory transient. Gaussian white noise is added to give different signal-to-noise ($SNR$) ratio,

\vspace{-6pt}

\begin{eqnarray}
SNR = 20 ~ log \frac{\beta}{\sqrt{2.\sigma}} ~ \textrm{[dB]}, \label{snoise}
\end{eqnarray}

\noindent in which $\beta$ is the amplitude of the original voltage signal; $\sigma$ is the standard deviation of the Gaussian noise; and dB means deciBel -- one tenth of Bel. 

Voltage data from power systems usually have an $SNR$ from 40 to 70dB. We evaluate the evolving classifier subject to 20, 40, and 60dB $SNR$ -- being 20dB the harshest stochastic scenario. Ten thousand voltage waveforms are generated:

\vspace{4pt}

Class 1: 2,000 waveforms without disturbances; 

Class 2: 2,000 waveforms with spikes; 

Class 3: 2,000 waveforms containing notching; 

Class 4: 2,000 waveforms with harmonics; 

Class 5: 2,000 waveforms with oscillatory transient.

\vspace{4pt}

Time windows based on constant time intervals between landmarks are considered. Windows of different lengths (1, 4 and 10 cycles of the fundamental) are assessed in different experiments. The greatest peaks and valleys of the fundamental voltage in a window are rescaled in the range $[0,1]$. A phase angle within [$-\pi$, $\pi$] is randomly assigned to the starting point of a waveform. Waveforms are subject to noise \eqref{snoise}.

\begin{figure}[!h]
	\begin{center}
		\vspace{5pt}
		{\includegraphics[scale=0.41]{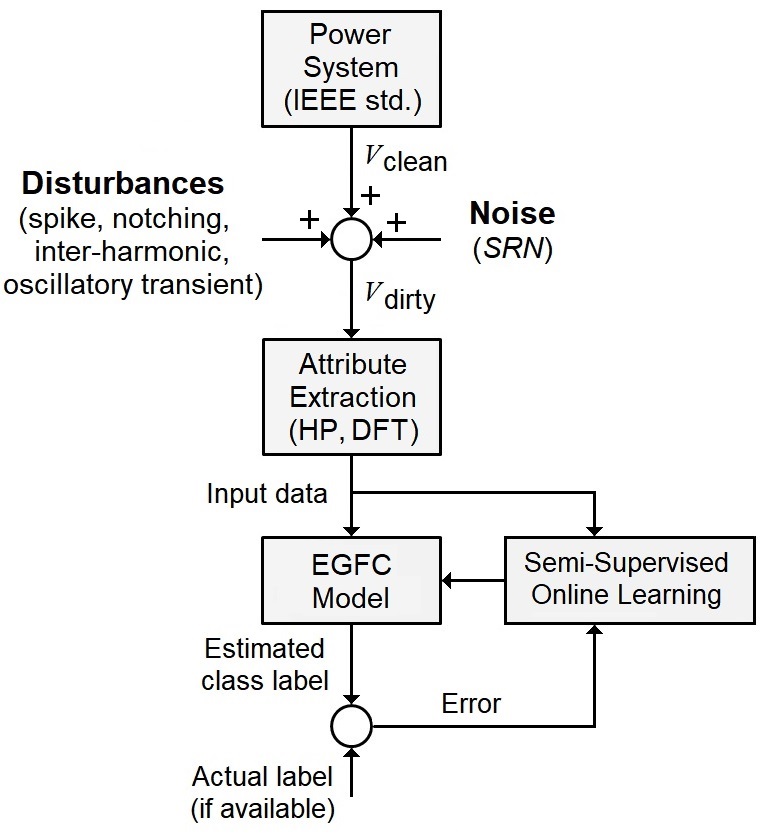}}
		\caption{Evolving Disturbance Detection and Classification System} \label{Fig1}
	\end{center}
\end{figure}

Voltage data within a window are fed to HP-DFT attribute extraction. Then, a vector of input data is formed and provided to the EGFC model. EGFC estimates a class, and then uses the input vector -- accompanied or not by a label -- to update its parameters and structure. For each window, this procedure is repeated. A general flowchart of the power quality monitoring and classification system is shown in Fig. \ref{Fig1}.

Four disturbance indicators compose an input vector $\textbf{x}^{[h]}$ of the EGFC model. They are

\vspace{4pt}

$x_1$: Amplitude of the fundamental (60Hz) voltage component, obtained by DFT over the data in a time window; 

$x_2$: Minimum value of the voltage cyclical component after HP decomposition over the data in a time window;

$x_3$: Maximum value of the voltage cyclical component after HP decomposition over the data in a time window;

$x_4$: Effective value of the voltage cyclical component after HP decomposition over the data in a time window.

\vspace{4pt}

\noindent The EGFC estimated output, $\hat{y}$, is a class $\hat{C} = \{1, 2, 3, 4, 5\}$. Constructing and updating an EGFC model is a fully online process based on partially-labeled data.

\subsection{Generating Disturbances}

Disturbances, viz., spike, notching, harmonics, and oscillatory transient (common power-system phenomena), are added to the fundamental voltage subject to an $SNR$.

Spikes or surges are fast, short-duration voltage transients caused by lightning strikes, power outages, short circuits, power transitions in large equipment, to mention some \cite{Johnson}. A spike usually lasts from 1 to 30$\mu s$, and may reach over 1,000V. For example, a motor when switched off can generate a spike of 1,000V. Spikes can degrade wiring insulation and destroy electronic devices. Some common-mode voltage spikes may not be detected by surge protection equipment.

\begin{figure}[h]
	\begin{center}
		\vspace{5pt}
		{\includegraphics[scale=0.25]{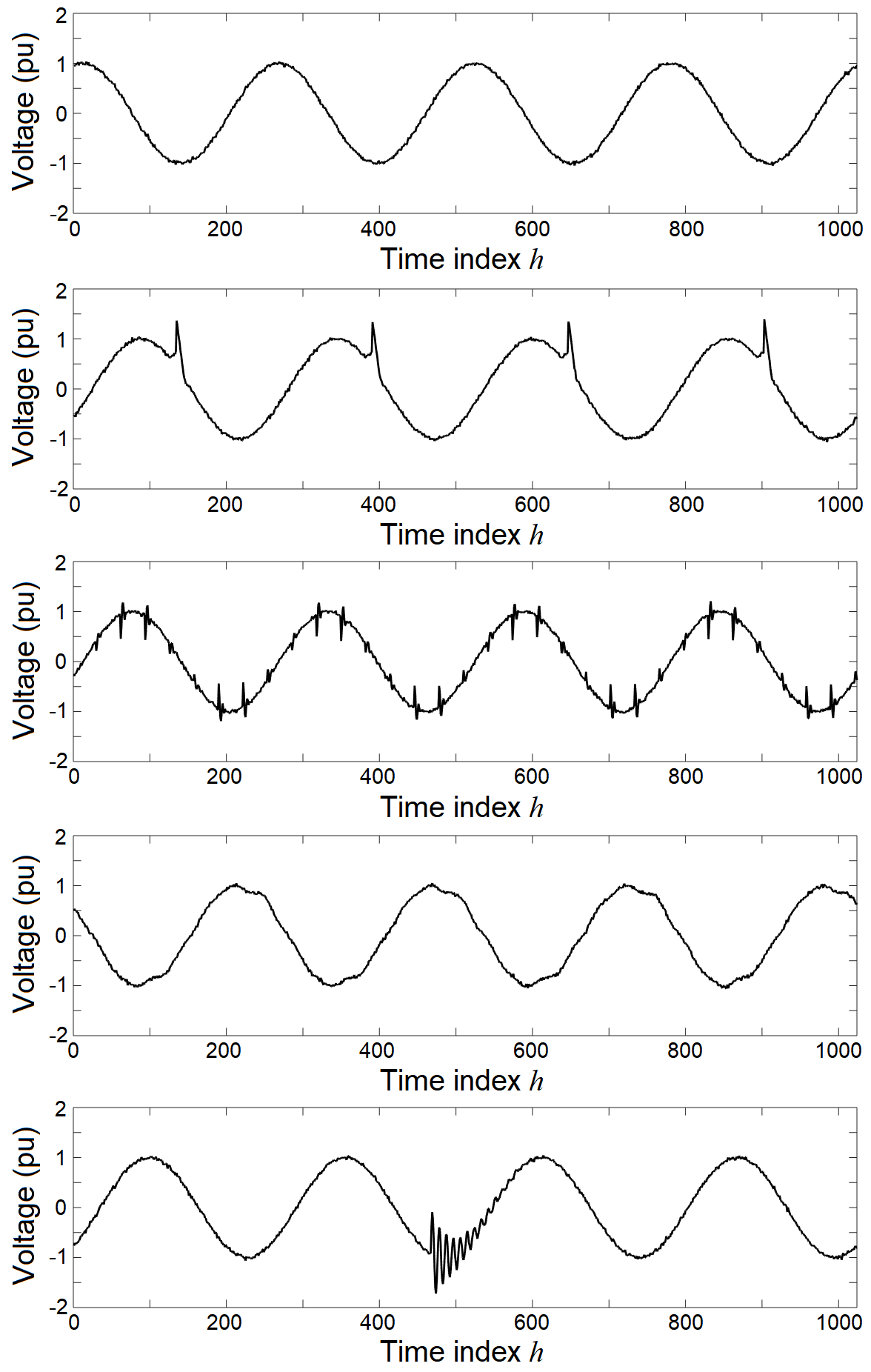}}
		\caption{Examples of 4-cycle voltage waveforms of each class, with $SNR = 30$. From top to bottom: no disturbance (class 1); spikes (class 2); notching (class 3); harmonics (class 4); oscillatory transient (class 5)} \label{Fig2}
	\end{center}
\end{figure}

To generate spike we choose a random starting point during the first voltage cycle. The spike peaks after 10 samples, and extinguishes after 20 samples. Its maximum amplitude is a random number in $[1, 1.5]$pu or $[-1.5, -1]$pu. The spike repeats in subsequent cycles for window lengths larger than one cycle, see example in Fig. \ref{Fig2}.

Notching is a periodic disturbance, a switching lasting less than 0.5 cycles. It is caused by the normal operation of electronic devices and three-phase converters. Notches occur when the current commutates from one phase to another. The severity of a notch is given by the source and isolating inductances of a converter, the magnitude of the current, and the point being monitored. The frequency components associated with notching can be quite high \cite{std519}. To generate notching we choose a random starting sample in [10, 40]. The disturbance extinguishes after 9 samples; it repeats 8 times per cycle, 23 samples after the previous occurrence. The maximum amplitude is $[-0.5, -0.05]$pu or $[0.05, 0.5]$pu, see Fig. \ref{Fig2}.

Harmonics are sinusoidal voltages with frequencies that are integer multiples of the fundamental. Distortion arises as current sources that inject harmonic currents into the power system cause nonlinear voltage drops across the system impedance. Harmonic currents result from the normal operation of nonlinear electronic devices and loads on the system. Harmonic distortion is a growing concern for customers and for the overall power system due to an increasing number of power electronics equipment \cite{std1159}. To generate harmonics, random values in $[0.008, 0.016]$pu, $[0.02, 0.04]$pu, $[0.005, 0.01]$pu, $[0.023, 0.046]$pu, $[0.003, 0.006]$pu, and $[0.02, 0.04]$pu are chosen, respectively, for the second to the seventh harmonic. The start point of each harmonic is independent one another and may have any phase angle in $[-\pi,\pi]$, see Fig. \ref{Fig2}.

\vspace{-5pt}

\vspace{-5pt}


Oscillatory transient is a sudden change of the voltage steady-state condition that includes rapid changes of positive and negative polarity values. Transients are almost always due to some type of switching event. Power electronic devices can produce oscillatory transients as a result of commutation and RLC snubber circuits \cite{std1159}. We choose a random sample in a time window as start point of a transient. The transient is an exponentially damped sinusoid whose start amplitude is in $[0.45, 1]$pu in relation to the fundamental component. Its frequency is a random value in $[1000,2500]$Hz. The damping coefficient is a random value in $[400, 1000]$, see Fig. \ref{Fig2}.

\subsection{Classification Accuracy}

Classification accuracy is computed recursively from

\vspace{-3pt}

\begin{eqnarray}
Acc(new) = \frac{h-1}{h} ~ Acc(old) + \frac{1}{h} ~ \tau,
\end{eqnarray}

\noindent in which $Acc \in [0,1]$; $\tau = 1$ if $\hat{C}^{[h]} = y^{[h]}$ (right estimate). Otherwise, $\tau = 0$ (wrong class estimate).

The average number of granules or rules over time, $c_{avg}$, is a measure of model concision. Recursively,

\vspace{-8pt}

\begin{eqnarray}
c_{avg}(new) = \frac{h-1}{h} ~ c_{avg}(old) + \frac{1}{h} ~ c^{[h]}.
\end{eqnarray}

\vspace{1pt}

\section{Results and Discussions} \label{experimental}

We evaluate the EGFC approach. No prior knowledge about the data and power system is assumed. Classification models are developed from scratch, based on data streams.

\subsection{Preliminary Results on Feature Extraction}

\begin{figure*}[!th]
	\begin{center}
		\vspace{5pt}
		{\includegraphics[scale=0.25]{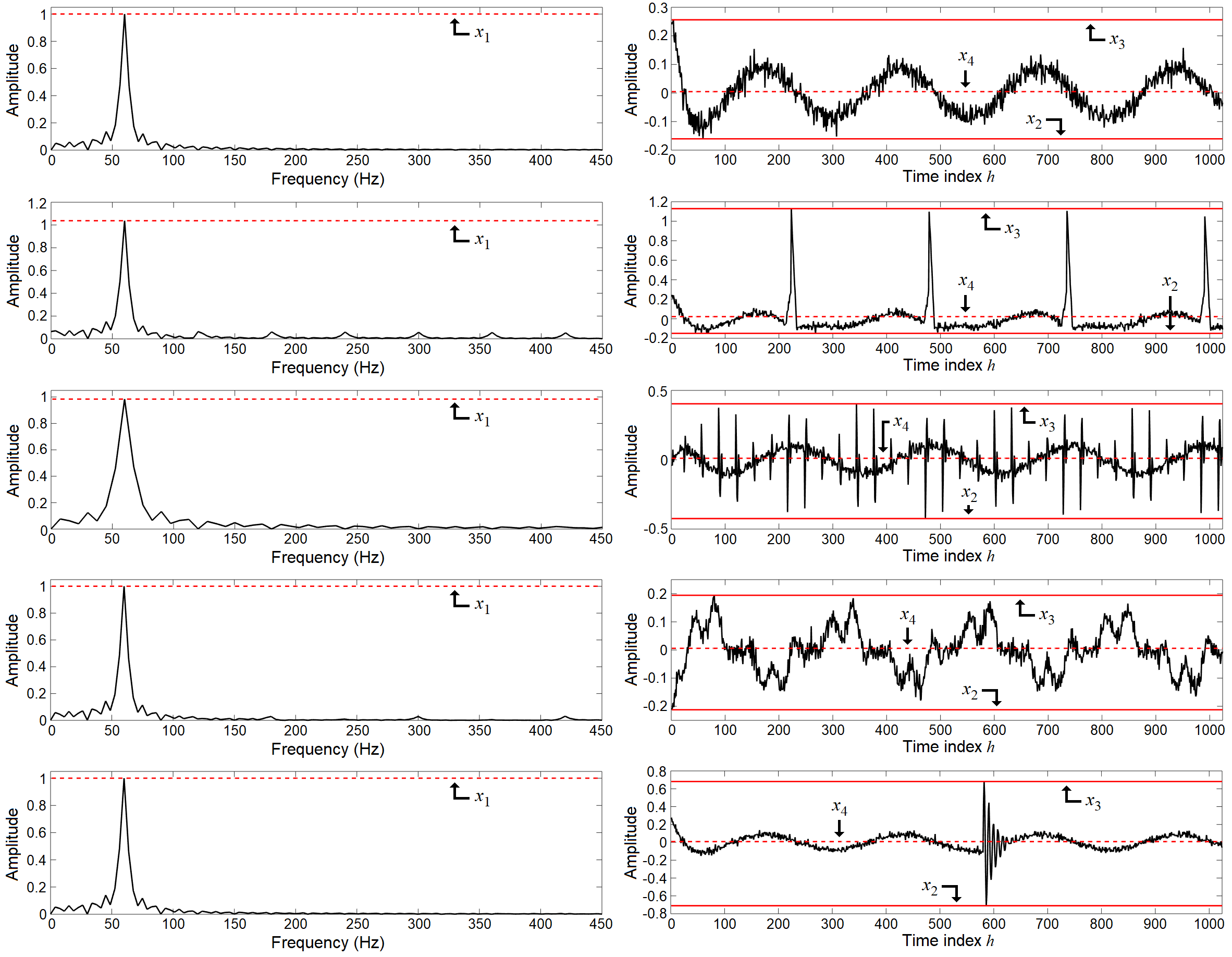}}
		\caption{Typical examples of attributes $\textbf{x} = [x_1 ~ x_2 ~ x_3 ~ x_4]$ extracted from 5 different voltage waveforms using DFT (left column) and HP filter (right column) considering 4-cycle time windows and an $SNR$ of 30. From top to bottom: row 1 - no disturbance (class 1); row 2 - spikes (class 2); row 3 - notching (class 3); row 4 - harmonics (class 4); row 5 - oscillatory transient (class 5) \label{Fig3}}
	\end{center}
\end{figure*}

We use DFT and HP filtering to extract four disturbance indicators from the voltage waveform. The DFT provides $x_1$, the amplitude of the fundamental component. The HP filter decomposes the original waveform into trend and cyclical components. We focus on the cyclical component to obtain $x_2$, $x_3$ and $x_4$, i.e., the minimum, maximum and effective values of the decomposed waveform. We achieved encouraging results using an HP smoothing coefficient of $256\times10^3$ as a great portion of noise is isolated from the fundamental. Figure \ref{Fig3} shows typical values of attributes, $\textbf{x} = [x_1 ~ x_2 ~ x_3 ~ x_4]$, extracted from waveforms of each disturbance class. The examples consider 4-cycle time windows and an $SNR$ of 30.

Notice in Fig. \ref{Fig3} that the value of $x_1$ changes slightly, up and down, respectively, in the spike and notching scenarios, which may help the evolving classifier to distinguish these classes. The HP cyclical component for the case without disturbance shows an initial transient such that the absolute values of $x_2$ and $x_3$ are significantly different one another, which helps the classifier to recognize this class. This phenomenon also happens for the spike scenario, with greater unbalance between the values of $x_2$ and $x_3$. Attribute $x_4$ tends to zero rapidly for the notching case. The absolute values of $x_2$ and $x_3$ in the harmonic scenario are similar. The same happens in the oscillatory transient case, but with higher individual amplitudes. Therefore, $\textbf{x}$ carries important subtleties.

\subsection{EGFC Results for Labeled Data Streams}

We look for an evolving classifier based on a data stream. The default meta-parameters are used (Sec. III-H). Table \ref{Tab1} shows the results averaged over 5 runs for 9 datasets extracted from voltage waveforms based on window lengths of 1, 4, and 10 cycles; and an $SNR$ of 20, 40, and 60dB. Each dataset consists of 10,000 4-attribute samples related to a target class $C^{[h]} \in \{1,2,3,4,5\}$. The classes mean `no disturbance', `spikes', `notching', `harmonics', and `oscillatory transient'.

Table \ref{Tab1} shows that the $SNR$ is irrelevant to the classifier performance using the set of attributes $\textbf{x}$ chosen. The accuracy can be relatively higher in noisier conditions, e.g. 20dB. This is an interesting feature of the proposed monitoring system. In contrast, a very small window length can degrade system performance significantly. The 4-cycle scenario seems more attractive than the 10-cycle one as the system is able to analyze a higher amount of windows at the price of a small reduction of the classification accuracy. As at least two 4-cycle windows are processed during a 10-cycle period, if the system provides wrong classification for the data extracted from the first window, it can still detect the disturbance class from the data of the other window. Therefore, in practice, the 92.79\%-accuracy 4-cycle-based EGFC system can be more efficient than the 94.24\%-accuracy 10-cycle-based one. The number of rules in the model structure over the learning steps, and the CPU time in a quad-core i7-8550U with 1.80GHz and 8GB of RAM are similar in all scenarios.

\begin{table}[h]
	\small \caption{EGFC Performance in Multiclass Classification of Power System Disturbances (99\% Confidence)}
	\vspace{-15pt}
	\begin{center}
		\resizebox{\columnwidth}{!}{
			\begin{tabular}{cc|ccc}
				\hline
				$SNR$ & Cycles & $Acc$(\%) & \# Rules & Time (s) \\
				\hline
				& 10 & $93.17 \pm 0.72$ & $9.15 \pm 1.73$ & $1.80 \pm 0.25$ \\
				60dB & 4 & $87.47 \pm 0.60$ & $9.52 \pm 1.73$ & $1.72 \pm 0.21$ \\
				& 1 & $64.41 \pm 0.27$ & $10.77 \pm 1.50$ & $1.97 \pm 0.23$ \\ 
				\hline                    
				& 10 & $92.98 \pm 1.30$ & $8.98 \pm 0.99$ & $1.69 \pm 0.10$ \\
				40dB & 4 & $88.33 \pm 0.74$ & $9.01 \pm 1.46$ & $1.67 \pm 0.12$ \\
				& 1 & $63.90 \pm 0.39$ & $10.58 \pm 0.60$ & $1.96 \pm 0.16$ \\
				\hline 
				& 10 & $94.24 \pm 0.21$ & $8.38 \pm 0.54$ & $1.59 \pm 0.08$ \\
				20dB & 4 & $92.79 \pm 1.07$ & $8.70 \pm 0.89$ & $1.65 \pm 0.12$ \\
				& 1 & $67.10 \pm 0.64$ & $9.69 \pm 1.36$ & $1.87 \pm 0.16$ \\
				\hline
			\end{tabular}
			\label{Tab1}}
	\end{center}
\end{table}

\vspace{-7pt}

\begin{figure}
	\begin{center}
		\vspace{5pt}
		{\includegraphics[scale=0.277]{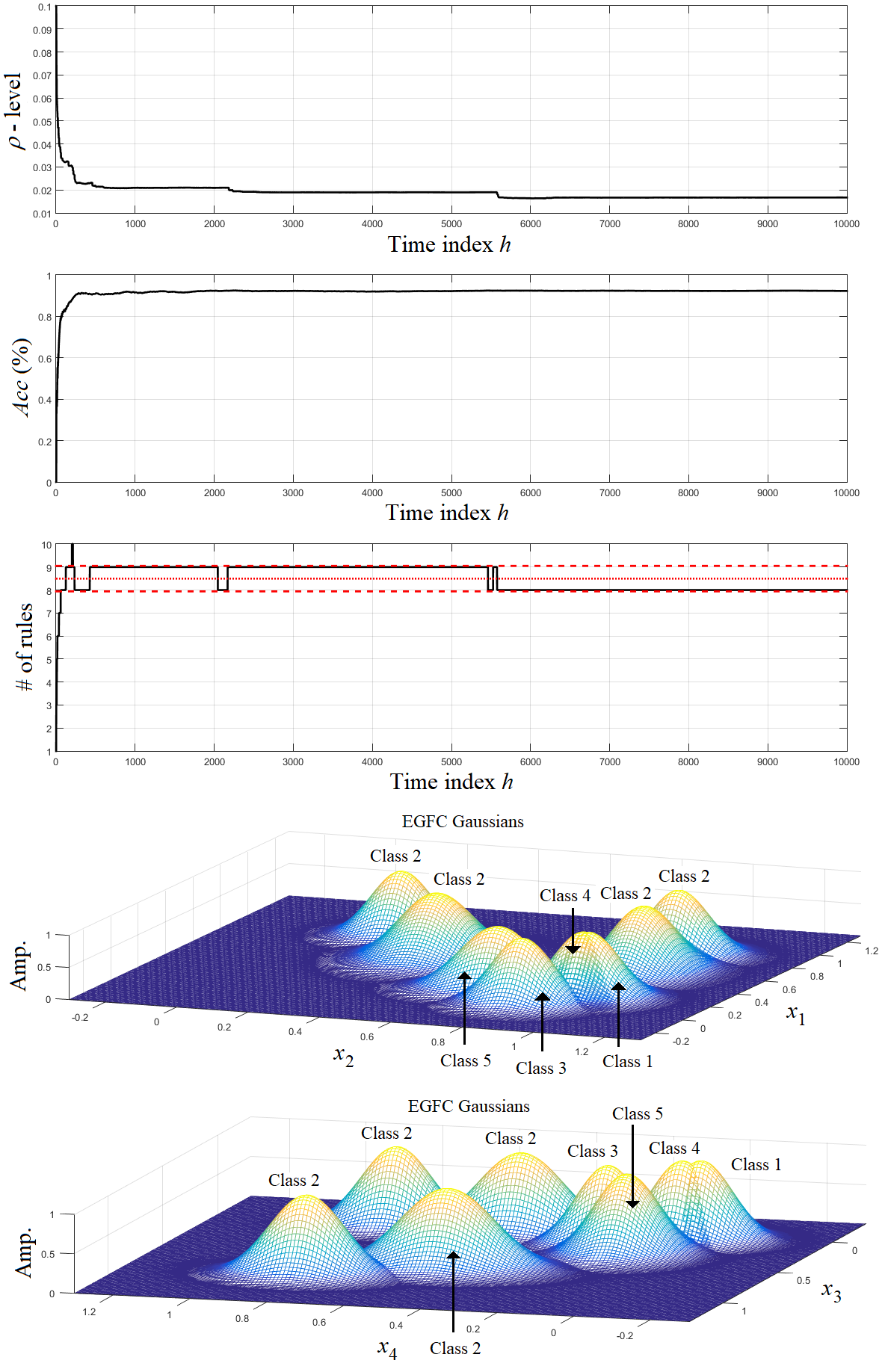}}
		\caption{Evolution of the $\rho$ level, EGFC accuracy, and number of rules. The bottom plots show the final shape of 4-dimensional Gaussian granules \label{Fig4}}
	\end{center}
\end{figure}
\begin{figure}
	\begin{center}
		\vspace{5pt}
		{\includegraphics[scale=0.62]{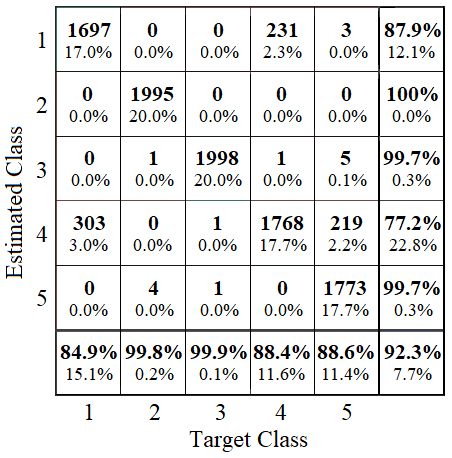}}
		\caption{Typical EGFC confusion matrix based on a 4-cycle time window, an $SNR$ of $20$dB, and labeled data stream \label{Fig5}}
	\end{center}
\end{figure}

Figure \ref{Fig4} shows a typical example of evolution of the $\rho$-level, accuracy, and number of EGFC rules. The final granules, at $h = 10,000$, are also illustrated. Class-2 data (spike disturbance) spread over a larger area, and require four granules to be represented, whereas the remaining classes require a single granule. Figure \ref{Fig5} shows the confusion matrix obtained. Class 1 (no disturbance) and Class 4 (harmonics), followed by Class 4 and Class 5 (transient), are those responsible for the 7.7\% overall error. Additional attributes should be considered in the future to address these particular hesitancies.

\subsection{EGFC Result in Semi-Supervised Online Scenario}

\begin{figure}[!ht]
	\begin{center}
		\vspace{5pt}
		{\includegraphics[scale=0.27]{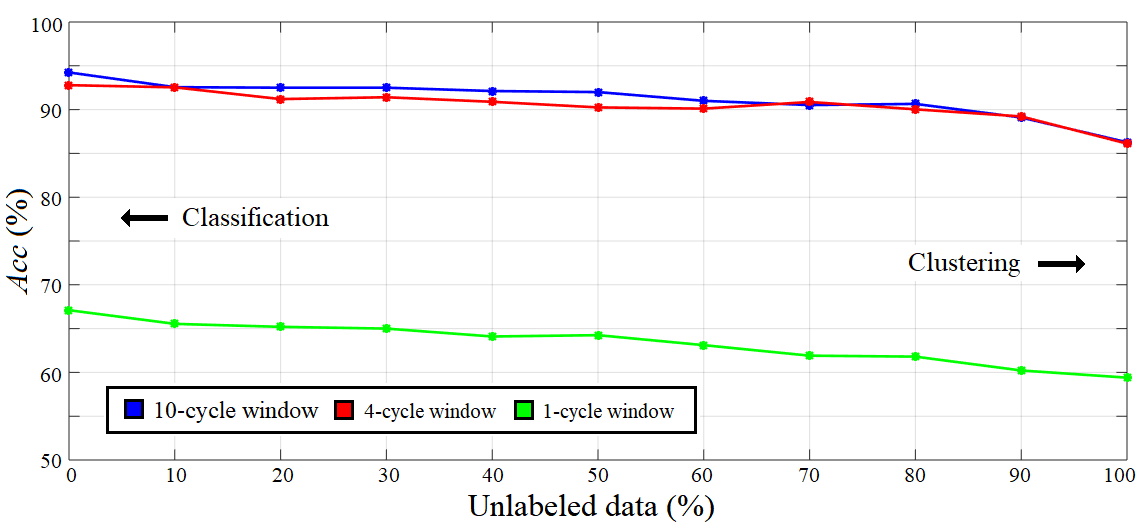}}
		\caption{EGFC performance using different proportions of unlabeled data} \label{Semi}
	\end{center}
\end{figure}

We changed the proportion of unlabeled data from 0\% to 100\% considering the harshest 20dB problem. Figure \ref{Semi} shows average EGFC results for 5 runs for each case. EGFC benefits of all information of the data stream, including that from unlabeled samples. Conventional and evolving classifiers that operate on a supervised basis by simply discarding unlabeled data cannot deal with small fractions of labeled data with reasonable accuracy (as shown by the right-side points of the graph). The left and right extremes of the plot indicate full supervision and non-supervision. In all cases the final result is a partition of data into classes. EGFC is not significantly affected by fractions of unlabeled data. For data extracted from 4-cycle windows, the performance of pure classification and clustering were $92.79\%$ and $86.12\%$, respectively.

\section{Conclusion} \label{conclusion}

We propose a hybrid attribute-extraction method combined with an evolving Gaussian fuzzy model for power quality disturbance detection and classification. A Hodrick-Prescott filter and the Discrete Fourier Transform applied over window of voltage data have provided attractive attributes for disturbance discrimination. Common types of disturbances, namely, spikes, notching, harmonics, and oscillatory transient were analyzed. Data generation agrees with the IEEE standard for power quality disturbances. A landmark window containing one, four, and ten voltage cycles as well as signal-to-noise ratio ranging from 20 to 60dB were evaluated. The evolving modeling approach, EGFC, has shown to be efficient for multi-class online classification. Its fuzzy rule-based structure, Gaussian membership functions, and granularity are updated over time driven by the data stream. Online model adaptation has shown to be essential to deal with time-varying systems, such as power systems subject to disturbance patterns.

Harmonics followed by oscillatory transient have shown to be more challenging to be distinguished compared to spikes and notching. The signal-to-noise ratio does not affect EGFC performance significantly. The use of 4 voltage cycles for attribute extraction provided an average EGFC accuracy of 92.8\% in the harshest 20dB noise case, and therefore was considered ideal. Changing the proportion of unlabeled data from $0\%$ to $100\%$ made the EGFC performance reduce from $92.8\%$ to $86.1\%$. Therefore, EGFC is applicable to and robust to clustering and classification. New types of disturbances can be studied in the future. The EGFC semi-supervised learning framework shall be analyzed considering synthetic examples of concept change and anomaly detection problems.


\end{document}